# DisMo: A Morphosyntactic, Disfluency and Multi-Word Unit Annotator
# An Evaluation on a Corpus of French Spontaneous and Read Speech

**George Christodoulides[1], Mathieu Avanzi[2], Jean-Philippe Goldman[3]**

[1] Centre Valibel, Institute for Language & Communication, University of Louvain
Place Blaise Pascal 1, B-1348 Louvain-la-Neuve, Belgium

[2] Formal Linguistics Laboratory, University Paris Diderot
Immeuble Olympes de Gouges, Rue Albert Einstein, FR-75013 Paris, France

[3] Linguistics Department, University of Geneva
Rue de Candolle 2, CH-1211 Geneva, Switzerland

E-mail: george@mycontent.gr, mathieu.avanzi@gmail.com, jean-philippe.goldman@unige.ch

**Abstract**

We present DisMo, a multi-level annotator for spoken language corpora that integrates part-of-speech tagging with basic disfluency detection and annotation, and multi-word unit recognition. DisMo is a hybrid system that uses a combination of lexical resources, rules, and statistical models based on Conditional Random Fields (CRF). In this paper, we present the first public version of DisMo for French. The system is trained and its performance evaluated on a 57k-token corpus, including different varieties of French spoken in three countries (Belgium, France and Switzerland). DisMo supports a multi-level annotation scheme, in which the tokenisation to minimal word units is complemented with multi-word unit groupings (each having associated POS tags), as well as separate levels for annotating disfluencies and discourse phenomena. We present the system's architecture, linguistic resources and its hierarchical tag-set. Results show that DisMo achieves a precision of 95% (finest tag-set) to 96.8% (coarse tag-set) in POS-tagging non-punctuated, sound-aligned transcriptions of spoken French, while also offering substantial possibilities for automated multi-level annotation.

## 1. Introduction

We present DisMo, a multi-level annotator for spoken corpora that integrates part-of-speech tagging with basic disfluency detection and annotation, and multi-word unit recognition. DisMo is a hybrid system that uses a combination of lexical resources, rules, and statistical models based on Conditional Random Fields (CRF). The system is trained and its performance evaluated on a 57k-token corpus of spoken French.

DisMo is designed to explicitly take into account the particular characteristics of spoken language. In the absence of punctuation, the annotator relies on prosodic features and discourse markers to identify discourse boundaries (cf. Leech, 1997; Mertens & Simon, 2013). Disfluencies, such as filled pauses, repetitions and false starts, affect up to 10% of tokens in natural conversation (Shriberg, 2001:154), while previous work has shown that part-of-speech tagging and downstream processing can be improved by detecting and marking these phenomena (e.g. Liu et al., 2006; Georgila, 2010). Furthermore, integrating multi-word expression identification can improve the performance of a POS tagger (e.g. Constant & Sigogne, 2011). While automatic boundary prediction, multi-word unit identification and disfluency detection have already been applied independently on spoken corpora (particularly in French), DisMo integrates these processing steps and encodes the interactions between them. The system's architecture is not tied to a particular language; however, the tag-set, lexical resources and statistical models have to be adapted to a specific language. In this paper, we present the first public version of DisMo for French, and the results of its evaluation.

This work builds upon an earlier version of the system (Christodoulides & Grosman 2012): the main improvements concern the processing of multi-word units and disfluencies, in addition to the use of a lager corpus for training and evaluation.

## 2. Presentation of DisMo

### 2.1 Input and Output

DisMo accepts several types of input: for a full analysis, an orthographic transcription aligned at the token level with the corresponding sound files is required. It is possible to use the system without the sound signal, in which case some of the prosodic features are ignored. It is also possible to use a transcription which is aligned at the utterance level only, in which case the resulting tokenisation is only approximately aligned. Annotating dialogues is also supported (either one file per speaker, or multiple speakers' tiers the same file along with a speaker identification tier).

The input formats may be a set of *Praat* (Boersma & Weenink, 2014) TextGrids, *TranscriberAG* (Barras et al., 1998), *ELAN* (Brugman & Russel, 2004), *Exmaralda Partitur* (Schmidt & Wörner, 2009), or tab-separated text files. DisMo may add its output as a set of annotation tiers in the above-mentioned formats (within the constraints of each format), and additionally supports outputting XML files, OpenDocument spreadsheets, or updating an SQL relational database in the *Praaline* (Christodoulides, 2014) format.

Figure 1: The multi-level annotation system of DisMo (upper three tiers from EasyAlign).

## 2.2 A multi-level Annotation Scheme

DisMo's output consists of six tiers: minimal tokens (`tok-min`), POS tag of minimal tokens (`pos-min`), multi-word units (`tok-mwu`), POS tag of multi-word units (`pos-mwu`), discourse markers and related phenomena annotation (`discourse`) and disfluency annotation (`disfluency`). Figure 1 shows sample output, in the format of a Praat TextGrid, highlighting the containment relationships between the three different levels: tiers `tok-min`, `pos-min` and `disfluency` are congruent; `tok-mwu` and `pos-mwu` are congruent and group minimal tokens into multi-word expressions; and `discourse` may independently group tokens in order to annotate discourse markers. In this figure, the tier '`transcription`' was the input to DisMo, '`spk2`' contains the utterances of the secondary speaker and tier '`speaker`' identifies the current speaker.

## 2.3 Annotation Process

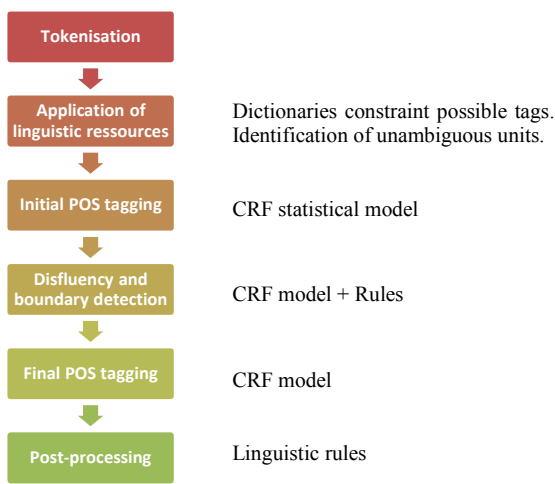

Figure 2: Cascade of processing steps

DisMo follows a cascade of annotation steps, in which each step refines the results of the previous ones.

Annotation modules operate on a shared data structure (the token list) that handles the various tags attributed or refined throughout the process, and the grouping of minimal tokens into multi-word expressions. (cf. Fig. 2). The following steps are applied to each corpus sample:
- Tokenisation.
- Application of linguistic resources: unambiguous tokens (including filled pauses and transcribed false starts), as well as potential discourse markers and multi-word units are identified.
- Preliminary part-of-speech annotation (CRF statistical model).
- Boundary and disfluency detection (combination of CRF models and rules).
- Final, combined part-of-speech and multi-word unit identification (CRF statistical model).
- Rule-based refining of the proposed tags.

DisMo is written in C++ and uses a series of open-source libraries. Dictionaries are stored as finite-state transducers using the Helsinki Finite-State Transducer Technology (HFST) library[1]. The Conditional Random Field (CRF) models are trained and applied using the CRF++ toolkit[2].

## 3. Language Resources

### 3.1 Corpus

In this paper we present the results of training and evaluating DisMo on a corpus of spoken French (Avanzi, 2014) created from PFC material (Durand et al. 2002, 2009). The corpus includes 12 regional varieties of French recorded in 3 different countries: 4 varieties spoken in Metropolitan France; 4 varieties spoken in Switzerland and 4 varieties spoken in Belgium. In total, there are 96 speakers in the corpus: For each of the 12 sites, 4 female and 4 male speakers, born and raised in the city they were recorded, were selected. The age of the speakers varies between 20 and 80. It is similar between the 12 groups of speakers (F (11, 95) = 0.360, n.s.), between male and female speakers (F (1, 95) = 0.82, n.s.) and between male and female speakers across the 12 groups (F (11, 95) = 0.133, n.s.).

---
[1] http://www.ling.helsinki.fi/kielitekniologia/tutkimus/hfst/
[2] http://crfpp.googlecode.com/svn/trunk/doc/index.html

The recordings consist in semi-directed socio-linguistic interviews, in which the informant has minimal interaction with the interviewer. In average, three minutes of spontaneous speech for each speaker are ortho-graphically transcribed and automatically aligned with the EasyAlign script (Goldman 2011) within Praat, which provides a 3-layer annotation in phones, syllables and words. All alignments were manually verified and corrected when necessary by inspecting both spectrogram and waveforms.

In total, the corpus is approximately 7 hour-long, and includes approximately 57k tokens. Table 1 presents the basic properties of this corpus. An expert annotator corrected the corpus POS tags and two of the authors verified the corrections. We used this corpus for training and evaluation using a 10-fold cross-validation method. A separate testing corpus of read speech (a text of 398 words, read by the same speakers) was also used.

| Region | City | Tokens | Time (min) |
|---|---|---|---|
| France | Paris | 4855 | 30 |
| | Lyon | 4448 | 27 |
| | Brécey | 5289 | 31 |
| | Ogéviller | 4773 | 30 |
| Belgium | Brussels | 5235 | 28 |
| | Liège | 4089 | 28 |
| | Tournai | 4744 | 28 |
| | Gembloux | 5749 | 28 |
| Switzerland | Geneva | 4752 | 26 |
| | Neuchâtel | 4306 | 27 |
| | Nyon | 4321 | 27 |
| | Martigny | 4424 | 28 |

Table 1: Corpus used for training and evaluation

### 3.2 Lexical Resources

Language-specific resources are used in the pre-processing stage and include:
- a set of tokeniser rules,
- a dictionary providing all the possible POS tags for each token, and
- a dictionary of potential multi-word units.

Applying these resources before the statistical annotation limits the search space. DisMo's language resources were compiled by merging publicly available dictionaries. For French, these include: DELA (Courtois et al., 1997) that contains simple forms and multi-word expressions and is distributed as part of the Unitex platform (Paumier, 2002), GLÀFF (Sajous et al., 2013) and manually-built lists of named entities. The dictionary POS tags were converted to the format supported by DisMo (cf. section 3.3).

## 4. Annotations and tag-sets

The part-of-speech (POS) tag set is based on a trade-off between the theoretical utility of each tag and the need for tagging precision. POS tags are organised in two main levels: a grammatical category (adjectives, adverbs, conjunctions, determiners, nouns, prefixes, pronouns, prepositions, verbs, interjections and foreign words) and a subcategory (type of adverbs, e.g. interrogative or gradation; type of determiner or pronoun, e.g. definite, demonstrative, possessive; type of noun, e.g. common, acronym or named entity; and verb mood and tense). A third level is used to indicate the syntactical function of the numeral in its context (e.g. a number that could be replaced by a noun will be tagged :nom). Auxiliary verbs are marked :aux at the third level. A fourth, "Extended-POS" level provides information about gender, number and person for verbs, nouns and adjectives; however this level is based on the dictionary entries and, as the current version of DisMo does perform a syntactical analysis (necessary for matching the constituents of the sentence), it may be ambiguous. The hierarchical system of POS tags allows querying the corpus at varying degrees of granularity. The POS tag-set is more precise and therefore automatically convertible to widely used French POS tag-sets, e.g. the POS tag-set of the French Treebank (Abeillé et al., 2003) or the TCOF corpus (Benzitoun et al., 2012).

Regarding disfluencies, DisMo uses a taxonomy based on Shriberg (2001). Simple disfluencies, i.e. affecting only one (minimal) token are filled pauses, hesitation-related lengthening, lexical false starts, and intra-word pauses. Structured disfluencies are analysed into three parts: the reparandum (disfluent), the interregnum (optional explicit editing terms), and the repair (fluent). They include repetitions, deletions, substitutions and insertions. Complex disfluencies are combinations of several simple and/or structured disfluencies. When simple disfluencies are identified at step (4) of processing (cf. Figure 2), they are excluded from the data submitted to final POS tagging. In this way, DisMo handles the circularity problem, i.e. the fact that disfluency identification improves the tagging, while disfluency detection is improved by the availability of POS tags.

Silent pauses are categorised as short or long (either on the basis of a user-defined threshold, or based on the statistical distribution of their lengths). Finally, a probabilistic model is used to identify potential discourse markers. Table 2 presents the full tag-sets for POS and disfluency annotations.

It should be noted that DisMo is customisable to different transcription conventions, for example regarding the method used to indicate false starts; symbols used for annotating paraverbal phenomena (e.g. coughing, laugh-ter); strings that should be ignored in the transcription, etc.

| | Part of Speech tag-set | | | | |
|---|---|---|---|---|---|
| Adjectives | ADJ | Adjective | | PRO:dem | Pronoun, demonstrative |
| Adverbs | ADV | Adverb | | PRO:ind | Pronoun, indefinite |
| | ADV:acr | Adverb | | PRO:int | Pronoun, interrogative |
| | ADV:int | Adverb, interrogative | | PRO:per:sjt | Pronoun, personal, subject |
| | ADV:neg | Adverb, negative | | PRO:per:objd | Pronoun, personal, direct object |
| Conjunctions | CON:coo | Conjunction, co-ordinating | Pronouns | PRO:per:obji | Pronoun, personal, indirect object |
| | CON:sub | Conjunction, subordinating | | PRO:pos | Pronoun, posessive |
| Determiners | DET:def | Determiner, definite article | | PRO:rel | Pronoun, relative |
| | DET:dem | Determiner, demonstrative | | PRO:ref | Pronoun, reflexive |
| | DET:ind | Determiner, indefinite | | PRO:per:ton | Pronoun, personal, clitic (stressable) |
| | DET:int | Determiner, interrogative | | Concatenated forms are annotated with tag1|tag2 | |
| | DET:exc | Determiner, exclamative | | VER:cond | Verb, conditional perfect |
| | DET:par | Determiner, partitive | | VER:cond:aux | Verb, conditional perfect, auxiliary |
| | DET:pos | Determiner, posessive | | VER:fut | Verb, future |
| Numerals | NUM:crd:det | Cardinal number, determiner | | VER:fut:aux | Verb, future, auxiliary |
| | NUM:crd:adj | Cardinal number, adjective | | VER:impe | Verb, imperative |
| | NUM:crd:pro | Cardinal number, pronoun | | VER:impf | Verb, imperfect |
| | NUM:crd:nom | Cardinal number, noun | | VER:impf:aux | Verb, imperfect, auxiliary |
| | NUM:ord:adj | Ordinal number, adjective | | VER:inf | Verb, infinitive |
| | NUM:ord:pro | Ordinal number, pronoun | | VER:inf:aux | Verb, infinitive, auxiliary |
| | NUM:ord:nom | Ordinal number, noun | | VER:ppas | Verb, past participle |
| Foreign | FRG | Foreign word | Verbs | VER:ppre | Verb, perfect participle |
| Interjections | ITJ | Interjection | | VER:ger | Verb, gerundive (only on MWU tier) |
| | ITJ:(category) | …specifying original POS | | VER:pres | Verb, present |
| | ITJ:ono | Onomatopoeia | | VER:pres:aux | Verb, present, auxiliary |
| Nouns | NOM:acr | Noun, acronym | | VER:pres:entatif | Verb, existential (*voilà, voici*) |
| | NOM:com | Noun, common | | VER:simp | Verb, simple past |
| | NOM:pro | Noun, proper | | VER:simp:aux | Verb, simple past, auxiliary |
| | NOM:pro:acr | Noun, proper | | VER:subi | Verb, subjunctive, imperfect |
| Prefixes | PFX | Prefix | | VER:subi:aux | Verb, subjunctive, imperfect, aux. |
| Prepositions | PRP | Preposition | | VER:subp | Verb, subjunctive, present |
| | PRP:det | Preposition + Determinant | | VER:subp:aux | Verb, subjunctive, present, auxiliary |
| | Disfluency Annotation tag-set | | | | |
| Simple disfluencies affect only 1 token. Structured disfluencies follow the pattern (reparandum) * < interregnum > repair. Complex disfluencies are a combination of several simple and structured ones. | | | Simple disfluencies | FIL | Filled pause |
| | | | | LEN | Hesitation-related lengthening |
| | | | | FST | Lexical false start |
| | | | | WDP | Pause within word |
| Structured disfluencies | REP | Repetition (one or more words) | Codes for disfluency structure | * | Interruption point |
| | DEL | Deletion | | -E | Explicit editing term |
| | SUB | Substitution, revision | | _ (underscore) | Repair part |
| | INS | Insertion | Complex | COM | Complex disfluency |

Table 2: The detailed hierarchical tag-set for French POS and disfluency annotation.

## 5. Evaluation

In order to evaluate the performance of DisMo, we used the 10-fold cross-validation methodology. The corpus was split into 10 "folds" that contain an approximately equal number of pause-separated units (PSUs) from each sub-corpus. We selected a high threshold for PSUs (500 ms), to ensure that these units would have been separated anyway by DisMo's boundary detection algorithm. One fold is set aside for testing while the remaining nine ones constitute the training corpus, and the process is repeated 10 times, for each fold. Table 3 summarises the results (averages over all 10 folds).

| | |
|---|---|
| Precision pos-min, level 1 | 96.8% |
| Precision pos-min, level 2 | 95.9% |
| Precision pos-min, entire tag | 95.0% |
| Disfluency detection precision | 74.5% |
| Disfluency detection recall | 55.4% |
| Disfluency classification precision | 68.7% |

Table 3: Evaluation results

## 6. Conclusion

In the speech community, there is a tacitly demonstrated need for shared annotated corpora, and an even greater need for freely accessible, user-friendly and robust annotation tools. In this context, the tool we present is in this paper is of great interest to researchers working on morphosyntactic, prosodic and discourse phenomena, and their interfaces. For example, it allows querying an annotated corpus, with a view to studying the relationship between prosody and syntax, in a more elaborate way than it is usually possible (i.e. without detailed annotation). Furthermore, to our knowledge, no robust systems exist for the automatic detection and annotation of disfluencies in French, despite some efforts to improve their automatic detection (e.g. Adda-Decker et al. 2004, Kalinli et al. 2009). The elaboration of such a system still represents a challenge for corpus linguistics. DisMo is a step in meeting this challenge and offers new perspectives for processing, studying and understanding these phenomena. Finally, DisMo is not limited to spoken corpora: the annotator can be applied to texts as well, especially in order to compare spoken and written language.

DisMo's algorithms are essentially language-independent: a version for English is already under development, as well as a generic version that can be trained on a user-supplied tag-set and corpus. DisMo is open source (GPL) software and is made freely available for non-commercial purposes. DisMo can be downloaded from the following address:

**www.corpusannotation.org/dismo**

## 7. Acknowledgements

Special thanks to Giulia Barreca and Julie Peuvergne (MoDyCo, University of Paris X Nanterre) for their detailed comments and suggestions that led to the version of the tag-set presented above.